\documentclass{article}

     \PassOptionsToPackage{numbers, sort&compress}{natbib}

     \usepackage[preprint]{neurips_2020}

\usepackage[utf8]{inputenc} 
\usepackage[T1]{fontenc}    
\usepackage{hyperref}       
\usepackage{url}            
\usepackage{booktabs}       
\usepackage{amsfonts}       
\usepackage{nicefrac}       
\usepackage{microtype}      

\usepackage{subcaption}
\usepackage{multirow}
\usepackage{graphicx}
\usepackage{tikz}
\usepackage{pgfplots}
\usetikzlibrary{calc}

\pgfplotsset{compat=1.11,
    /pgfplots/ybar legend/.style={
    /pgfplots/legend image code/.code={%
       \draw[##1,/tikz/.cd,yshift=-0.25em]
        (0cm,0cm) rectangle (3pt,0.8em);},
   },
}

\pgfplotsset{
    /pgfplots/layers/Bowpark/.define layer set={
        axis background,axis grid,main,axis ticks,axis lines,axis tick labels,
        axis descriptions,axis foreground
    }{/pgfplots/layers/standard},
}

\definecolor{pa1}{RGB}{27,161,226}
\definecolor{pa2}{RGB}{15,153,51}
\definecolor{pa3}{RGB}{216,0,115}
\definecolor{pa4}{RGB}{240,150,9}
\definecolor{pa5}{RGB}{162,0,255}
\definecolor{pa6}{RGB}{0,171,169}
\definecolor{pa7}{RGB}{0,200,200}

\title{Post-Training BatchNorm Recalibration}

\author{%
  Gil Shomron\\
  Faculty of Electrical Engineering\\
  Technion --- Israel Institute of Technology\\
  \texttt{gilsho@campus.technion.ac.il} \\
  \And
  Uri Weiser\\
  Faculty of Electrical Engineering\\
  Technion --- Israel Institute of Technology\\
  \texttt{uri.weiser@ee.technion.ac.il} \\
}

\begin{document}

\maketitle

\begin{abstract}
We revisit non-blocking simultaneous multithreading (NB-SMT) introduced previously by \citet{shomron2020non}.
NB-SMT trades accuracy for performance by occasionally "squeezing" more than one thread into a shared multiply-and-accumulate (MAC) unit.
However, the method of accommodating more than one thread in a shared MAC unit may contribute noise to the computations, thereby changing the internal statistics of the model.
We show that substantial model performance can be recouped by post-training recalibration of the batch normalization layers' running mean and running variance statistics, given the presence of NB-SMT.
\end{abstract}

\section{Introduction}

Simultaneous multithreading (SMT) is a well-known technique used mainly in general purpose processors (GPPs) to mitigate hardware underutilization \citep{yamamoto1994performance, yamamoto1995increasing, tullsen1995simultaneous, eggers1997simultaneous}.
SMT works by dispatching instructions from multiple threads to shared execution units, thereby increasing efficiency and achieving better performance.
It does so in an opportunistic manner, that is, if instruction dependencies are met and if the needed resources are available, the instruction will be executed; otherwise, the instruction will be blocked and will wait in a reservation station.

Inspired by conventional SMT, \citet{shomron2020non} propose non-blocking SMT (NB-SMT) designated for deep neural networks (DNNs).
NB-SMT mitigates hardware underutiliation of DNNs caused by unstructured sparsity.
In contrast to GPPs, which are obliged to keep program semantics so as to produce consistent results, DNNs are able to cope with, for example, pruning of activations and weights \citep{han2015learning, shomron2020thanks} and with quantization \citep{shkolnik2020robust, banner2019post} without retraining.
NB-SMT exploits DNN resiliency to avoid backpressure \citep{shomron2019smt} by "squeezing" more than one thread into the shared resource by temporarily reducing the threads' numerical precision.

In this short note, we revisit NB-SMT and show that some of the model performance can be recouped after training (i.e., post-training) by regathering the batch normalization \citep{ioffe2015batch} (or just BatchNorm), layers' running mean and running variance statistics.
The model accuracy degradation we observe is a result of dynamic reduced numerical representation and the deviation of internal statistics that accompanies it.
While actual training (i.e., gradient computations and weight updates) is required to compensate for the former, the latter can be compensated for in an online unsupervised manner by recalibrating the BatchNorm statistics, given, in our case, NB-SMT.
Recall that, traditionally, these statistics are gathered during training and are considered constant during inference, or they may even be folded into the weights of adjacent layers.

The importance of recalibrating the BatchNorm layers due to an external corrupted dataset or internal perturbations in activations and/or weights has been discussed in other works as well.
\citet{schneider2020improving} show how BatchNorm recalibration can improve model robustness of vision models to image corruptions (e.g., blurring and compression artifacts); \citet{tsai2020calibrated} propose to recalibrate the BatchNorm layers in the scenario of noise in analog accelerators; \citet{shomron2020thanks} recalibrate the BatchNorm layers to redeem some of the accuracy degradation due to zero-valued activation mispredictions; and \citet{hubara2020improving}, as well as \citet{shomron2020non}, suggest post-quantization BatchNorm recalibration.
Other works \citep{ioffe2017batch, summers2020four, singh2019evalnorm, guo2018double} tackle the problem of mitigating BatchNorm training and inference discrepancy due to relatively small batch sizes.
Interestingly, \citet{summers2020four} also mention that data augmentation differences between the two modes of operation may also be a source of discrepancy.
From an analytical point of view, \citet{sun2019hybrid} show that retuning BatchNorm statistics has the potential to significantly reduce the quantization error --- the expected 2-norm quantization error at a BatchNorm layer output increases linearly with the variance of the quantization error when the original BatchNorm mean and variance statistics are kept, but increases only sub-linearly when the BatchNorm statistics are recouped.





\section{Method}

\citet{shomron2020non} showed that a single 8-bit multiply-and-accumulate (MAC) unit may be shared across a number of CNN threads, where a thread comprises an activation vector and a weight vector that are to be multiplied and accumulated.
Many times, due to the inherent sparsity of CNN models, no more than one thread requires the MAC unit.
If, however, more than one thread does require the shared MAC unit, the threads' precision is reduced momentarily, so that the MAC unit can accommodate all threads in parallel.
This "squeezing" operation is what makes NB-SMT non-blocking and adds noise during inference.

We would like to compensate for the corruption in the internal statistics of the model due to NB-SMT by regathering the BatchNorm statistics.
To do so, we let the BatchNorm layers collect the new exponential moving average of the mean and variance on a subset of the training set with NB-SMT.
Notice that collecting these statistics does not require gradient computations and does not modify the BatchNorms' $\gamma$ and $\beta$ parameters.

%

\section{Experiments}

\begin{table}[t]
\caption{Top-1 ImageNet accuracy with different CNN models given 2-threaded (2T) and 4-threaded (4T) NB-SMT precision reduction mechanism with and without BatchNorm recalibration.}
\label{tbl:results}

\centering
\resizebox{\textwidth}{!}{%
\begin{tabular}{@{}clcc|cc|cc@{}}
\toprule
\multicolumn{1}{l}{} & \multicolumn{1}{c}{\textbf{}} & \textbf{} & \textbf{} & \multicolumn{2}{c|}{\textbf{2T}} & \multicolumn{2}{c}{\textbf{4T}} \\
\textbf{Source} & \multicolumn{1}{c|}{\textbf{Model}} & \textbf{FP32} & \textbf{A8W8} & \textbf{w\textbackslash{}o} & \textbf{w\textbackslash} & \textbf{w\textbackslash{}o} & \textbf{w\textbackslash} \\ \midrule
\multirow{4}{*}{PyTorch} & \multicolumn{1}{l|}{ResNet-18} & 69.76 & 69.70 & 68.49 (-1.27) & 68.86 (-0.90) & 63.74 (-6.02) & 67.90 (-1.86) \\
 & \multicolumn{1}{l|}{ResNet-50} & 76.15 & 76.24 & 75.10 (-1.05) & 75.82 (-0.33) & 70.44 (-5.71) & 73.49 (-2.66) \\
 & \multicolumn{1}{l|}{GoogLeNet}             & 69.78 & 69.63 & 67.45 (-2.33) & 67.79 (-1.99) & 59.84 (-9.94) & 65.83 (-3.95) \\
 & \multicolumn{1}{l|}{GoogLeNet$^{\dagger}$} & 69.78 & 69.63 & 69.25 (-0.53) & 69.18 (-0.60) & 63.55 (-6.23) & 66.75 (-3.03) \\
 & \multicolumn{1}{l|}{DenseNet-121} & 74.65 & 74.66 & 74.05 (-0.60) & 74.36 (-0.29) & 71.94 (-2.71) & 73.57 (-1.08) \\ \midrule
\multirow{2}{*}{MLPerf} & \multicolumn{1}{l|}{ResNet-50} & 76.46 & 76.42 & 73.75 (-2.71) & 76.07 (-0.39) & 67.62 (-8.84) & 73.88 (-2.58) \\
 & \multicolumn{1}{l|}{MobileNet-v1$^{\ddag}$} & 71.68 & 71.41 & 70.03 (-1.65) & 70.20 (-1.48) & 46.09 (-25.6) & 67.80 (-3.88) \\ \bottomrule
 \multicolumn{8}{l}{\vspace{-8px}} \\
 \multicolumn{8}{l}{\footnotesize $^{\dagger}$ a single high MSE layer is not accelerated.} \\
 \multicolumn{8}{l}{\footnotesize $^{\ddag}$ depthwise convolution layers are not accelerated.}
\end{tabular}%
}

\end{table}

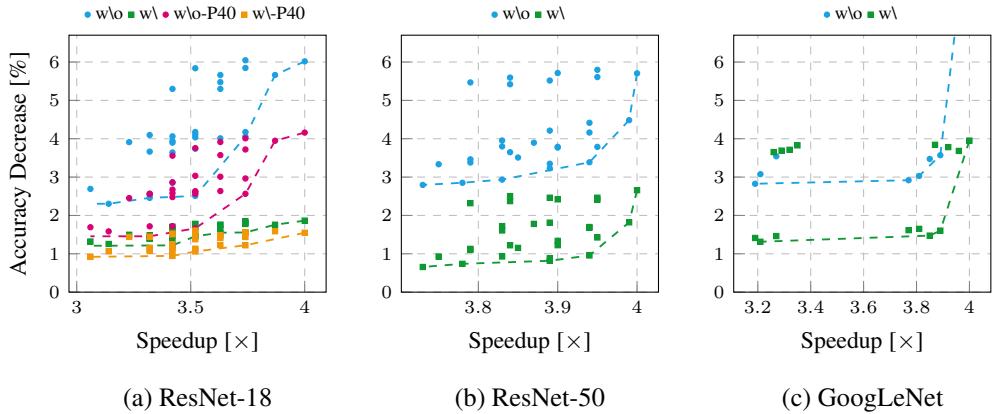
\begin{figure}[t] 
    \centering

		\begin{tikzpicture}
		
		\begin{axis}[
			name=plot1,
			title style={at={(0.5,-0.4)}, anchor=north}, title={(a) ResNet-18},
			ylabel={Accuracy Decrease [\%]}, xlabel={Speedup [$\times$]},
			width=5cm, height=5cm, 
			xtick pos=left, ytick pos=left, ymin={0}, ytick={0, 1, 2, 3, 4, 5, 6}, ymax={6.7},
			xlabel near ticks, ylabel near ticks,
			xmajorgrids, ymajorgrids, major grid style={dashed},
			x tick label style = {font=\scriptsize},
			y tick label style = {font=\scriptsize},
			legend style={font=\scriptsize, legend columns=4, at={(0.03, 1.08)}, anchor=west, draw=none},
			label style = {font=\footnotesize},
			set layers=Bowpark
			]
			
			\addplot[only marks, mark=*, mark size=1.0pt, color=pa1]  file {plotdata/rn18_pareto_clean_p0.dat};
			\addplot[only marks, mark=square*, mark size=1.0pt, color=pa2]  file {plotdata/rn18_pareto_bn_p0.dat};
			\addplot[only marks, mark=*, mark size=1.0pt, color=pa3]  file {plotdata/rn18_pareto_clean_p40.dat};
			\addplot[only marks, mark=square*, mark size=1.0pt, color=pa4]  file {plotdata/rn18_pareto_bn_p40.dat};	
			
			\addplot[color=pa1, line width=0.7pt, dashed]  file {plotdata/rn18_pareto_clean_p0_front.dat};
			\addplot[color=pa2, line width=0.7pt, dashed]  file {plotdata/rn18_pareto_bn_p0_front.dat};
			\addplot[color=pa3, line width=0.7pt, dashed]  file {plotdata/rn18_pareto_clean_p40_front.dat};
			\addplot[color=pa4, line width=0.7pt, dashed]  file {plotdata/rn18_pareto_bn_p40_front.dat};
			
			\legend{w\textbackslash{}o, w\textbackslash{}, w\textbackslash{}o-P40, w\textbackslash{}-P40};
		\end{axis}

		\begin{axis}[
			name=plot2, at={($(plot1.east)+(1cm,0)$)}, anchor=west,
			title style={at={(0.5,-0.4)}, anchor=north}, title={(b) ResNet-50},
			xlabel={Speedup [$\times$]},
			width=5cm, height=5cm, 
			xtick pos=left, ytick pos=left, ymin={0}, ytick={0, 1, 2, 3, 4, 5, 6}, ymax={6.7},
			xlabel near ticks, ylabel near ticks,
			xmajorgrids, ymajorgrids, major grid style={dashed},
			x tick label style = {font=\scriptsize},
			y tick label style = {font=\scriptsize},
			legend style={font=\scriptsize, legend columns=2, at={(0.32, 1.08)}, anchor=west, draw=none},
			label style = {font=\footnotesize},
			set layers=Bowpark
			]
			
			\addplot[only marks, mark=*, mark size=1.0pt, color=pa1]  file {plotdata/rn50_pareto_clean.dat};
			\addplot[only marks, mark=square*, mark size=1.0pt, color=pa2]  file {plotdata/rn50_pareto_bn.dat};
			
			\addplot[color=pa1, line width=0.7pt, dashed]  file {plotdata/rn50_pareto_clean_front.dat};
			\addplot[color=pa2, line width=0.7pt, dashed]  file {plotdata/rn50_pareto_bn_front.dat};
			
			\legend{w\textbackslash{}o, w\textbackslash{}};
		\end{axis}

		\begin{axis}[
			name=plot3, at={($(plot2.east)+(1cm,0)$)}, anchor=west,
			title style={at={(0.5,-0.4)}, anchor=north}, title={(c) GoogLeNet},
			xlabel={Speedup [$\times$]},
			width=5cm, height=5cm, 
			xtick pos=left, ytick pos=left, ymin={0}, ytick={0, 1, 2, 3, 4, 5, 6}, ymax={6.7},
			xlabel near ticks, ylabel near ticks,
			xmajorgrids, ymajorgrids, major grid style={dashed},
			x tick label style = {font=\scriptsize},
			y tick label style = {font=\scriptsize},
			legend style={font=\scriptsize, legend columns=2, at={(0.32, 1.08)}, anchor=west, draw=none},
			label style = {font=\footnotesize},
			set layers=Bowpark
			]
			
			\addplot[only marks, mark=*, mark size=1.0pt, color=pa1]  file {plotdata/gl_pareto_clean_t1.dat};
			\addplot[only marks, mark=square*, mark size=1.0pt, color=pa2]  file {plotdata/gl_pareto_bn_t1.dat};	
			
			\addplot[color=pa1, line width=0.7pt, dashed]  file {plotdata/gl_pareto_clean_t1_front.dat};
			\addplot[color=pa2, line width=0.7pt, dashed]  file {plotdata/gl_pareto_bn_t1_front.dat};
			
			\legend{w\textbackslash{}o, w\textbackslash{}};
		\end{axis}
		
		\end{tikzpicture}

	\caption{ImageNet top-1 accuracy decrease versus speedup w\textbackslash{}o BatchNorm recalibration and w\textbackslash{} BatchNorm recalibration.
			 To achieve a range of speedups (and accuracies), layers were executed with either 2 threads (2T) or 4 threads (4T).
			 (a) ResNet-18: 4T with some 2T and 40\% sparsity (P40); (b) ResNet-50: 4T with some 2T; (c) GoogLeNet: 4T with some 1T.}
	\label{fig:speedup_sweep}
\end{figure}

Throughout this section, we evaluate CNN accuracy results given SySMT --- an NB-SMT-enabled output-stationary systolic array \citep{shomron2020non}.
The convolution layers are transformed to fit the systolic array \citep{chetlur2014cudnn}.
We do not consider the first convolution layer and the fully connected layers, which we leave intact.
In addition, none of the results presented in this section consider reordering, as opposed to some of the results presented by \citet{shomron2020non}.
As for the quantization process, models are quantized with a simple 8-bit uniform min-max quantization, using unsigned per-layer quantization for activations and symmetric signed per-channel quantization for weights \citep{krishnamoorthi2018quantizing}.
The post-training quantization is performed according to statistics gathered on 2K images picked randomly from the training set.


With BatchNorm recalibration, we let the BatchNorm layers gather mean and variance statistics from the portion of the training set during inference using a 2-threaded (2T) and 4-threaded (4T) SySMT. By doing so, the BatchNorm layers can capture the noise induced by NB-SMT and align the distributions accordingly.
We may consider this process to be offline, since (1) it is performed on the training set (even though the training set labels are unnecessary); and (2) after it is completed the BatchNorm can be folded into an adjacent layer (usually convolution layers).

In Table~\ref{tbl:results}, we present NB-SMT results with SySMT for various image classification CNN architectures --- ResNet-18 and ResNet-50 \citep{he2016deep}, GoogLeNet \citep{szegedy2015going}, DenseNet-121 \citep{huang2017densely}, and MobileNet-v1 \citep{howard2017mobilenets} --- with the ILSVRC-2012 dataset \citep{russakovsky2015imagenet}, PyTorch vision model parameters \citep{paszke2019pytorch}, and MLPerf model parameters \citep{reddi2020mlperf}.
Clearly, by recalibrating BatchNorm statistics we achieve superior results across the board, besides a slight decrease with GoogLeNet$^{\dagger}$.

We also experiment with different SySMT operating points in which (1) part of the model is set to run with reduced acceleration to restore some accuracy; and (2) the model weights are pruned.
We conduct a simple unstructured weight pruning in an iterative method in which the weights are pruned according to their magnitude and the entire model is then retrained \citep{han2015learning}; specifically, we choose 40\% pruning of ResNet-18 weights.
We present our results in Fig.~\ref{fig:speedup_sweep}.
Each dot represents a measurement in which the acceleration of some of the layers is decreased, thereby trading performance for accuracy.
The question which layers are best to decelerate, and how much to decelerate them to achieve overall high speedup and high accuracy, is still an open question; as we can see, some layer choices present inferior accuracy results compared with others, when considering the same speedup, for example.
It is also clear from Fig.~\ref{fig:speedup_sweep}, as from Table~\ref{tbl:results} results, that by recalibrating the BatchNorm some accuracy can be regained.
In addition, Fig.~\ref{fig:speedup_sweep}(a) shows that additional unstructured weight sparsity can further increase model accuracy on top of the BN recalibration due to fewer NB-SMT thread collisions.

\section{Conclusion}
In this short paper, we provide a quick rundown of recent works that describe the importance of post-training BatchNorm recalibration, given DNN internal or external noise.
In that spirit, we revisit non-blocking simultaneous multithreading \citep{shomron2020non} and demonstrate how significant accuracy degradation can be recouped.

\bibliographystyle{abbrvnat}
\bibliography{refs}

\end{document}